\documentclass{article}
\usepackage[letterpaper, margin=0.5in]{geometry}

\usepackage{graphicx}
\usepackage{times}
\usepackage{algorithm}
\usepackage{algorithmic}
\usepackage{amsmath}
\usepackage{amsfonts}

\usepackage{authblk}
\usepackage[
  backend=biber,
  style=numeric,
]{biblatex}
\twocolumn
\title{CFR-ICL: Cascade-Forward Refinement with Iterative Click Loss for Interactive Image Segmentation}
\author[1]{Shoukun Sun}
\author[1]{Min Xian}
\author[2]{Fei Xu}
\author[2]{Luca Capriotti}
\author[2]{Tiankai Yao}
\affil[1]{\small Machine Intelligence and Data Analytics Lab, Department of Computer Science, University of Idaho}
\affil[2]{\small Idaho National Laboratory}

\date{
    \{ssun,mxian\}@uidaho.edu, \{fei.xu,luca.capriotti,tiankai.yao\}@inl.gov
}

\begin{document}

\maketitle

\begin{abstract}
The click-based interactive segmentation aims to extract the object of interest from an image with the guidance of user clicks. Recent work has achieved great overall performance by employing feedback from the output. However, in most state-of-the-art approaches, 1) the inference stage involves inflexible heuristic rules and requires a separate refinement model, and 2) the number of user clicks and model performance cannot be balanced. To address the challenges, we propose a click-based and mask-guided interactive image segmentation framework containing three novel components: Cascade-Forward Refinement (CFR), Iterative Click Loss (ICL), and SUEM image augmentation. The CFR offers a unified inference framework to generate segmentation results in a coarse-to-fine manner. The proposed ICL allows model training to improve segmentation and reduce user interactions simultaneously. The proposed SUEM augmentation is a comprehensive way to create large and diverse training sets for interactive image segmentation. Extensive experiments demonstrate the state-of-the-art performance of the proposed approach on five public datasets. Remarkably, our model reduces by 33.2\%, and 15.5\% the number of clicks required to surpass an IoU of 0.95 in the previous state-of-the-art approach on the Berkeley and DAVIS sets, respectively.
\end{abstract}

\section{Introduction}
Interactive image segmentation extracts object(s) of interest from images with user input. It is essential for enabling the broader application of deep learning-based solutions in real-world scenarios. Deep neural networks (DNNs) often require extensive annotated data, which could be extremely expensive and time-consuming to create due to high labor costs. Interactive segmentation offers a more cost-effective solution for generating large-scale labeled datasets.

Interactions in interactive segmentation include scribbles \cite{xianNeutroConnectednessCut2016a}, boxes \cite{lempitskyImageSegmentationBounding2009}, and clicks \cite{xianEISegEffectiveInteractive2016a}. This work focuses on the click-based approach, whereby the user provides positive and negative clicks on the image to identify foreground and background regions, respectively. Earlier click-based methods were developed using image processing techniques, such as the connectedness-based approach described in \cite{xianEISegEffectiveInteractive2016a}. However, subsequent advancements in the field led to the development of deep learning-based methods, beginning with \cite{xuDeepInteractiveObject2016}, which resulted in significant improvements in segmentation performance. Recently, more deep learning-based methods have been introduced, including \cite{jangInteractiveImageSegmentation2019,sofiiukFbrsRethinkingBackpropagating2020,sofiiukRevivingIterativeTraining2022,chenFocalClickPracticalInteractive2022}, which have further improved the efficiency of interactive segmentation.

Many interactive segmentation methods \cite{jangInteractiveImageSegmentation2019,sofiiukFbrsRethinkingBackpropagating2020,chenFocalClickPracticalInteractive2022,sofiiukRevivingIterativeTraining2022,liuSimpleClickInteractiveImage2022} define the loss functions based on the difference between the predicted segmentation masks and ground truths, and have no explicit way to optimize the number of user clicks. Some studies \cite{chenFocalClickPracticalInteractive2022} use a refinement of initial predictions to enhance the quality of outputs. These techniques necessitate additional trainable network modules to identify and fine-grind the local areas in the predicted masks, which increases the difficulty of training. In this paper, we put forward an Iterative Click Loss (ICL) that imposes penalties on instances with a large number of clicks. We present a novel strategy for refinement during inference time, named Cascade-Forward Refinement (CFR). The CFR is designed to improve segmentation details during the inference process, without the need for an additional network. Also, we disclose an effective image augmentation technique - SUEM Copy-Paste (C\&P). This technique is an extension of the Simple Copy-Paste \cite{ghiasiSimpleCopypasteStrong2021} method, specifically designed for interactive image segmentation tasks. The contributions of this work are summarized below.

\begin{itemize}
    \item To the best of our knowledge, the proposed Iterative Click Loss is the first loss that encodes the number of clicks to train a model for interactive segmentation; and it offers a novel approach to define preference on models with fewer user clicks.

    \item The proposed Cascade-Forward Refinement enhances the segmentation quality during inference in a simple and unified framework and can be applied to other iterative mask-guided interactive segmentation models.

    \item In the proposed SUEM augmentation, we propose a set of four copy-paste augmentation approaches that greatly improve the diversity and increase the size of training  sets in practical settings.
\end{itemize}

\begin{figure}[!th]
    \centering
    \includegraphics[width=1\columnwidth]{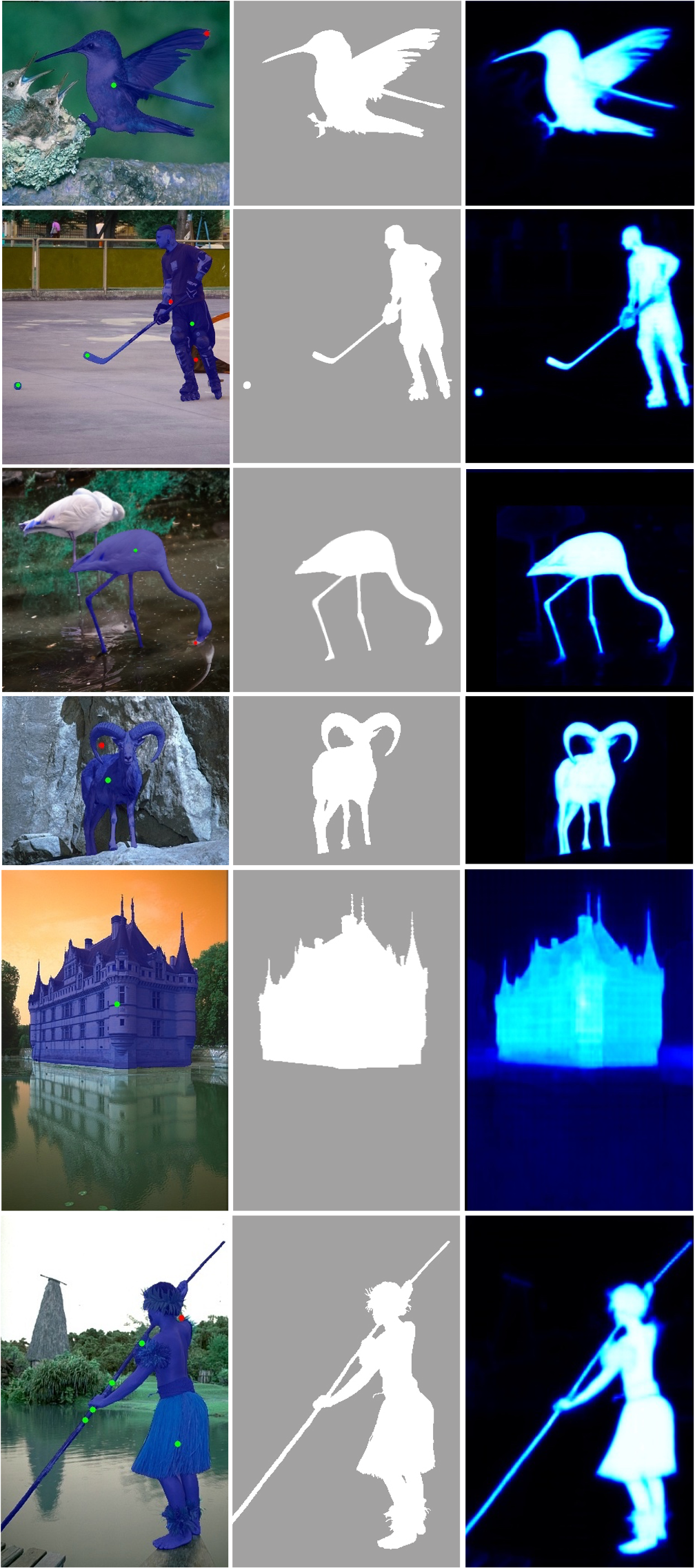}
    \caption{Examples of segmentation results that exceed an IoU of 0.95. The first column shows images with clicks (green for the foreground and red for the background) and segmentation masks from the proposed approach (blue). The second column shows the ground truth. The third column shows probability maps of the proposed approach. These raw images are from the Berkeley \cite{martinDatabaseHumanSegmented2001} and DAVIS \cite{perazziBenchmarkDatasetEvaluation2016} sets.}
    \label{fig:vis}
\end{figure}

\begin{figure*}[t]
    \centering
    \includegraphics[width=0.9\textwidth]{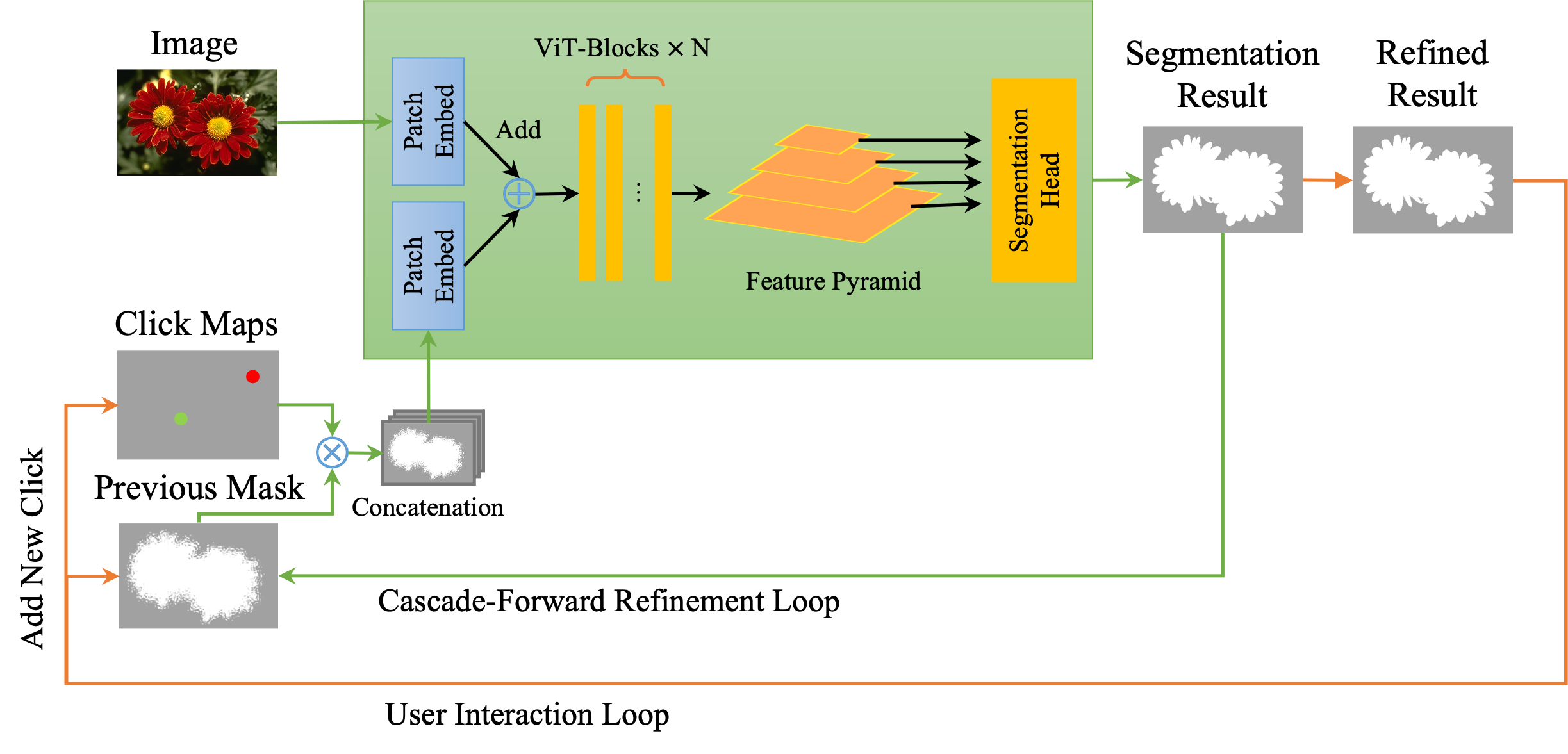}
    \caption{Overview of iterative mask-guided interactive segmentation integrated with Cascade-Forward Refinement. The orange colored lines represent the user interaction loop (outer loop). The green colored line represents the Refinement loop (inner loop). The black colored lines are shared processes for both loops. New clicks are added by the user in the user interaction loop. In the CFR loop, the previous mask is iteratively optimized with clicks.}
    \label{fig:flowchart}
\end{figure*}

\begin{figure*}[t]
    \centering
    \includegraphics[width=1\textwidth]{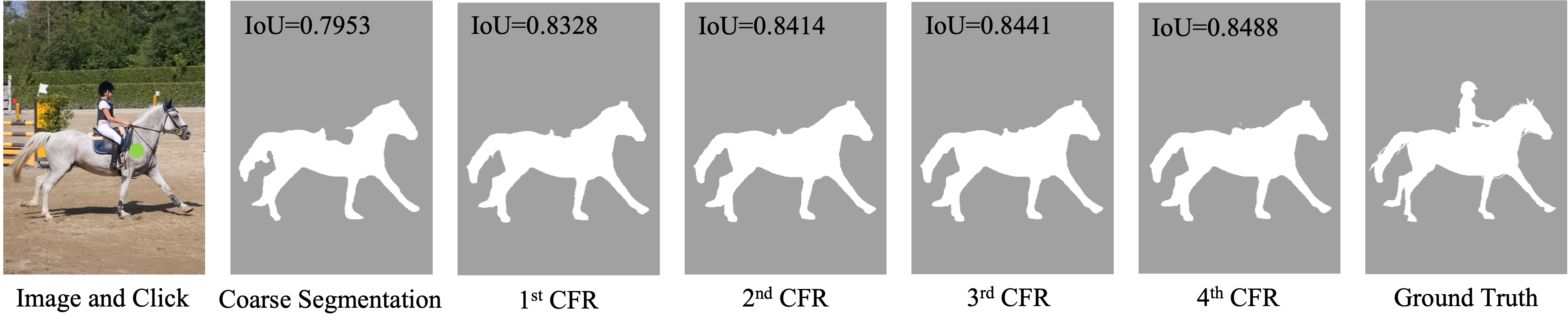}
    \caption{Sample results of Cascade-Forward Refinement.}
    \label{fig:cf}
\end{figure*}

\section{Related Work}

\textbf{Interactive segmentation}. Prior to the widespread adoption of deep learning techniques, interactive segmentation was primarily achieved through image processing-based methods, such as GrabCut \cite{rotherGrabCutInteractiveForeground2004}, NC-Cut \cite{xianNeutroConnectednessCut2016a}, and EISeg \cite{xianEISegEffectiveInteractive2016a}. With the emergence of deep learning, deep learning-based interactive segmentation models have obtained increasing popularity. The DIOS model \cite{xuDeepInteractiveObject2016} encoded the background and foreground user clicks into two distance maps and concatenated the image with the two maps as input. The BRS \cite{jangInteractiveImageSegmentation2019} and f-BRS \cite{sofiiukFbrsRethinkingBackpropagating2020} formulated the task as an inference-time online optimization problem. BRS optimizes the image or distance maps during inference to improve the segmentation results for given user clicks, while f-BRS optimizes the intermediate layer weights to achieve the same goal and speed up the inference. FCA-Net \cite{linInteractiveImageSegmentation2020} built a first-click attention module that enhances the importance of the first click as it usually plays a critical role to identify the location of the main body. UCP-Net \cite{dupontUCPNetUnstructuredContour2021} proposed a novel contour-based approach that asks the user to provide clicks on the contour of objects. 

\textbf{Iterative mask-guided interactive segmentation}. An iterative sampling strategy has been proposed by \cite{mahadevanIterativelyTrainedInteractive2018} which samples a single point from the center of a misclassified area iteratively. These iteratively generated points are used in training to boost performance. RITM \cite{sofiiukRevivingIterativeTraining2022} adopted and modified the iterative sampling strategy during training and iteratively uses the previous output as model input to achieve higher segmentation quality. The iterative mask-guided model only involves a simple feedforward process that is more computationally efficient than the inference-time optimization approaches such as BRS and f-BRS. More recently, FocalClick \cite{chenFocalClickPracticalInteractive2022} proposed a coarse-to-fine pipeline with a SegFormer \cite{xieSegFormerSimpleEfficient2021} backbone and achieved state-of-the-art results. It utilizes a heuristic strategy to determine local regions that possibly contain errors and uses a local refinement network to update these regions. SimpleClick \cite{liuSimpleClickInteractiveImage2022} greatly improved performance by adopting the Plain Vision Transformer (Plain ViT), which was pre-trained with MAE \cite{heMaskedAutoencodersAre2022}, as the backbone of the RITM approach. The advanced ViT network architecture has significantly benefited interactive segmentation. Our work builds upon the RITM and SimpleClick approaches and further explores the nature of interactive segmentation.

\textbf{Segment Anything Model}. The Segment Anything Model (SAM) \cite{kirillovSegmentAnything2023} represents an advanced development in image segmentation. Building on the foundations of interactive segmentation, SAM facilitates varied user interactions. The concept of a 'promptable segmentation task' underlies SAM's design, enabling the generation of valid segmentation masks in response to a variety of prompts including text, masks, clicks, and boxes. By integrating a heavyweight image encoder with a flexible prompt encoder and a fast mask decoder, SAM is capable of predicting multiple object masks, even in situations where the prompt may be ambiguous. SAM has the potential to function like a click-based interactive segmentation model when solely points and masks are utilized as input. To gauge its effectiveness on click-based interaction, we compare the SAM model with other leading click-based interactive segmentation models.

\textbf{Image augmentation}. Deep learning applications typically require large amounts of data to achieve optimal model performance, and the diversity of the training samples is crucial. Therefore, it is essential to develop efficient data augmentation techniques to enhance data efficiency. The previous study \cite{ghiasiSimpleCopypasteStrong2021} has shown that a simple copy-paste strategy can serve as a powerful data augmentation for instance-level segmentation tasks. We develop more specific image augmentation approaches for adapting interactive segmentation tasks.

\begin{figure*}[t]
\centering
\includegraphics[width=0.9\textwidth]{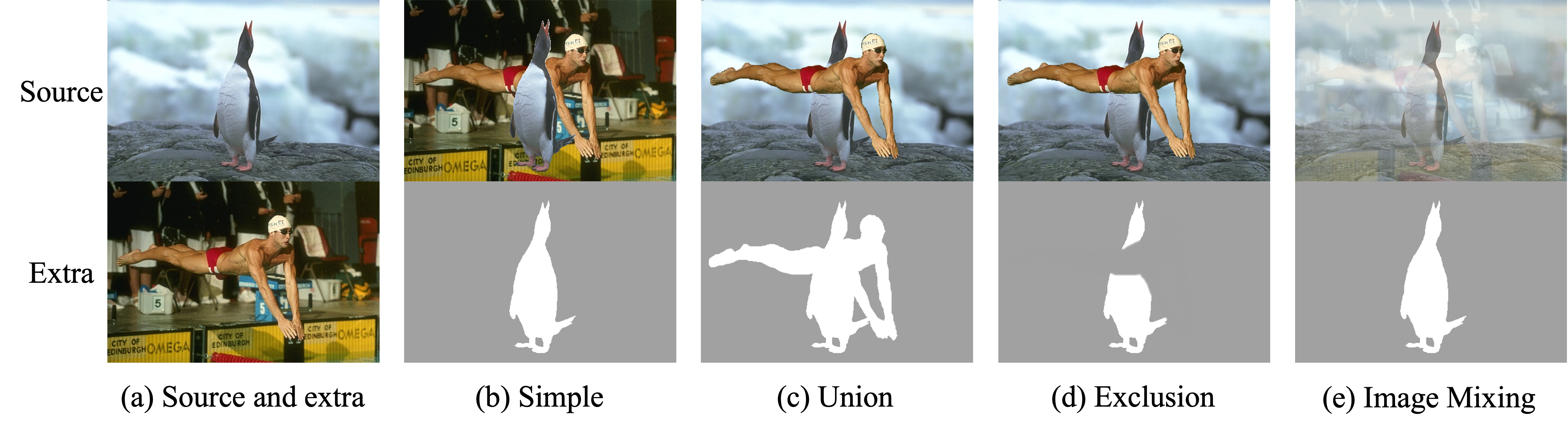}
\caption{Illustration of copy-paste modes.}
\label{fig:copypaste}
\end{figure*}

\section{Proposed Method}

We build a new iterative mask-guided framework that includes 1) a novel training loss that encodes the number of clicks into the loss term and defines the model preference of fewer clicks; 2) an inference-time refinement scheme without the need for extra modules;  3) an effective image augmentation designed for the interactive segmentation scenario.

\subsection{Iterative Click Loss}
Sofiiuk et al. \cite{sofiiukFbrsRethinkingBackpropagating2020} encoded all generated user clicks for each image into two disk maps and inputted them into a segmentation network during training. The training process has no differences from conventional deep neural networks for image segmentation.  In \cite{mahadevanIterativelyTrainedInteractive2018,sofiiukRevivingIterativeTraining2022}, researchers adopted a hybrid strategy that combined randomly sampled clicks with iteratively generated clicks to generate input maps. However, the trained models 1) may need more user interactions during inference, and 2) have no effective way to balance the number of inputted clicks and segmentation performance.

To overcome the challenges, we propose the Iterative Click Loss (ICL) approach which embeds the number of clicks during training. An initial set of randomly sampled clicks, denoted as $P^0$, is generated from the ground truth to forward the model and obtain an initial output $Y^{0}$. Clicks are generated iteratively (one click at a time) by sampling from the misclassified regions in $Y^0$, and the newly produced click is combined with $P^0$ to form a new sequence $P^1$. This process is repeated to generate the whole click sequence $P^t$. The click-sampling strategy can be formulated as

\begin{equation}\label{eq:itersample}
\begin{aligned}
P^0 &= \text{Random sample from } \mathbb{Y} \\
Y^0 &= f(X, P^0, \mathbf{0}) \\
P^1 &= P^0 \cup S(\mathbb{Y}, Y^0) \\
Y^1 &= f(X, P^1, Y^0) \\
&... \\
P^t &= P^{t-1} \cup S(\mathbb{Y}, Y^{t-1}) \\
Y^t &= f(X, P^t, Y^{t-1}),
\end{aligned}
\end{equation}

\noindent where $\mathbf{0}$ is the zero-initialized mask, $\mathbb{Y}$ is the ground truth, and $S$ is a sampling function \cite{sofiiukRevivingIterativeTraining2022} that proposes a single click among the misclassified areas in the output mask. A conventional total loss function \cite{sofiiukRevivingIterativeTraining2022} is defined by 

\begin{equation}\label{eq:oldloss}
L = \mathbb{L}(Y^t, \mathbb{Y}).
\end{equation}

\noindent where $\mathbb{L}$ is the Normalized Focal Loss \cite{sofiiukAdaptisAdaptiveInstance2019}. Note that in Eq. \ref{eq:oldloss}, a sequence of click sets $[P^0, P^1, ..., P^t]$ and a sequence of outputs $[Y^0, Y^1, ..., Y^{t-1}]$ are used to generate segmentation mask $Y^t$; however, only the final output is applied to calculate the Normalized Focal Loss and update the model parameters. In the proposed ICL approach,  a new loss function is built to accumulate the weighted losses of the generated mask sequence. 

\begin{equation}
\begin{aligned}
{L_{ICL}} &= \sum^t_{i=1} \beta_i \mathbb{L}(Y^i, \mathbb{Y}),
\end{aligned}
\end{equation}

\noindent where $\beta_i$ is used to control the weight of each term. Each click produces one loss term in the above equation, and minimizing the loss improves the segmentation performance and reduces the number of clicks simultaneously. By increasing the weights of the loss term for more clicks (larger $i$), the model is incentivized to use fewer clicks to achieve accurate segmentation. ICL offers a novel approach to define preference on interactive segmentation models with fewer user clicks.

\begin{table*}[th]
\centering
\begin{tabular}{r|cc|cc|cc|cc|cc}
\hline
&
\multicolumn{2}{c}{GrabCut} & 
\multicolumn{2}{c}{Berkeley} &
\multicolumn{2}{c}{DAVIS} &
\multicolumn{2}{c}{Pascal VOC} &
\multicolumn{2}{c}{SBD} \\
NoC@ & 90 & 95 & 90 & 95 & 90 & 95 & 90 & 95 & 90 & 95 \\
\hline
Models \\
\hline
\hline
SimpleClick & 1.54          & 2.16          & 2.46          & 6.71          & 5.48          & 12.23          & 2.81          & 3.75          & 5.24          & 11.23          \\
ICL          & 1.50          & 2.00          & 2.34          & 6.48          & 5.44          & 11.86          & \textbf{2.62} & \textbf{3.58} & \textbf{5.05} & \textbf{11.03} \\
SUEM C\&P         & \textbf{1.44} & \textbf{1.74} & \textbf{1.97} & \textbf{5.08} & \textbf{5.28} & \textbf{10.69} & 2.68          & 3.59          & 5.33          & 11.48          \\
\hline
\end{tabular}
\caption{\label{tab:conv} The Effectiveness of ICL and SUEM C\&P.}
\end{table*}

\subsection{Cascade-Forward Refinement}
Similar coarse-to-fine pipelines \cite{chenFocalClickPracticalInteractive2022} were proposed to incorporate a local refinement module to improve the details of local regions after the initial coarse segmentation of the entire image. These coarse-to-fine strategies demonstrate impressive overall performance; however, the approaches 1) depend on heuristic rules to select image regions for further refinement and 2) have to train two individual deep learning models independently.

We introduce the Cascade-Forward Refinement (CFR)-based inference strategy, which enables iterative refinement of the segmentation results without needing two models. The proposed CFR has two inference loops, i.e., the outer loop generates coarse segmentation masks using incremental user interactions, and the inner loop refines the masks by forwarding the segmentation model multiple times with the same input image and user clicks.

Let $f$ denote a deep neural network for segmentation, $Y^t$ be the output of the $t$-th step, and the $X$ be the input image. The sequence of user clicks at the $t$-th step is denoted as $P^t=\{(u_k,v_k,l_k)\}_{k=1}^{t}$, where $(u_k, v_k)$ represents the coordinates of a user click and $l_k \in \{0, 1\}$ denotes its label (0 for background and 1 for foreground). The model's inputs are raw images, maps generated by clicks, and previous segmentation masks. We define the CFR as 

\begin{equation}\label{eq:y0}
    \begin{aligned}
        Y^t_{0} = f(X,P^{t},Y^{t-1}_{n}), \text{and}
    \end{aligned}
    \end{equation}
    \begin{equation}\label{eq:yt}
    \begin{aligned}
        Y^t_{i} = f(X,P^t,Y^t_{i-1}), i \in \{1, 2, 3, ..., n\},
    \end{aligned}
\end{equation}

\noindent where $Y^t_{0}$ denotes the output of the $t$-th coarse segmentation  step (outer loop), and $Y^{t-1}_{n}$ is the last refined mask at step $t-1$. Eq. \ref{eq:yt} defines the refinement results at the $i$-th step in the inner loop, and $n$ defines the number of refinement steps which could be defined as a fixed number or determined adaptively. The Cascade-Forward Refinement approach iteratively updates the coarse segmentation masks and refinement masks using Eqs. \ref{eq:y0} and \ref{eq:yt}. No additional user clicks are required during the refinement process (inner loop), and the segmentation mask is continuously refined to provide higher-quality input. Figure \ref{fig:flowchart} illustrates the overview pipeline of the CFR and Figure \ref{fig:cf} shows a series of CFR refined results.

\textbf{Fixed-step CFR and Adaptive CFR Inference.} A fixed-step CFR applies $n$ times refinement for each user click. This refinement is employed during inference to enhance the quality of the output and can be integrated into any iterative mask-based model without the need for model modification. In addition to the fixed-step approach, we propose and validate an adaptive CFR (A-CFR) scheme. It counts the number of altered pixels between $Y^t_{i}$ and $Y^t_{i-1}$ and terminates the inner loop when the number of changed pixels falls below a specified threshold, or the maximum step is reached. We use CFR-$n$ and A-CFR-$n$ to denote fixed $n$ step CFR and adaptive CFR with maximum $n$ step, respectively.

\subsection{SUEM Copy-paste}

To generate large and diverse datasets for interactive segmentation scenarios, we proposed a comprehensive image augmentation method, namely SUEM Copy-paste (C\&P), that consists of four C\&P modes, Simple C\&P \cite{ghiasiSimpleCopypasteStrong2021}, Union C\&P, Exclusion C\&P, and the image Mixing.

The underlying principle of the C\&P method involves inserting randomly selected objects from one image into another. In the context of interactive segmentation, we refer to the object of interest and its corresponding image as the \textit{source object} and \textit{source image}, respectively. Conversely, the object and its corresponding image selected randomly from the training set are denoted as the \textit{extra object} and \textit{extra image}.

\textbf{Simple C\&P Mode.} The simple copy-paste mode involves inserting a source object into a randomly selected extra image, with the mask of the source object serving as the ground truth. Figure \ref{fig:copypaste} (b) provides a visual representation of this mode.

\textbf{Union C\&P Mode.} Interactive segmentation typically involves identifying a target object that comprises multiple objects, such as a man embracing a child. To simulate such a scenario, we employ the union copy-paste mode, where an extra object is pasted into the source image, and the ground truth is determined as the union of their respective masks. Figure \ref{fig:copypaste} (c) depicts the resulting image in this mode.

\textbf{Exclusion C\&P Mode.} Another scenario that frequently arises is that the object of interest is obstructed by another object, such as a person standing behind a pole. To address this issue, we introduce the exclusion copy-paste mode, where an extra object is copied into the source image, and the mask of the source object is utilized as the ground truth, excluding the mask of the extra object. Figure \ref{fig:copypaste} (d) provides a visual representation of this mode.

\textbf{Image Mixing Mode.} The approach of image-mixing involves blending a source image with an extra image and utilizing the mask of the source object as the ground truth. The image-mixing mode is depicted in Figure \ref{fig:copypaste} (e)

The above strategies are combined to generate training images in our experiments.

\begin{table}[th]
\centering
\begin{tabular}{c|c|c|c}
\hline
           & RITM           & EMC-Click           & Ours            \\
\hline
GrabCut    & 2.48           & \textbf{1.94}           & 1.96 \\
\hline
Berkeley   & 5.41           & 6.92           & \textbf{5.05}    \\
\hline
DAVIS      & 11.52          & 11.26          & \textbf{10.85}   \\
\hline
Pascal VOC & -              & \textbf{3.44}              & 3.57                       \\
\hline
SBD        & 12.00 & \textbf{11.68} & 12.02            \\
\hline
\end{tabular}
\caption{\label{tab:ritm} NoC@95 of RITM, EMC-Click \cite{du_efficient_2023} and proposed approach. HRNet32 backbone and training set C+L are used.}
\end{table}

\begin{table*}[th]
\centering
\begin{tabular}{r|cc|cc|cc|cc|cc}
\hline
&
\multicolumn{2}{c}{GrabCut} &
\multicolumn{2}{c}{Berkeley} &
\multicolumn{2}{c}{DAVIS} &
\multicolumn{2}{c}{Pascal VOC} &
\multicolumn{2}{c}{SBD} \\
\hline
NoC@ & 90 & 95 & 90 & 95 & 90 & 95 & 90 & 95 & 90 & 95 \\
\hline
Inference \\
\hline
\hline
StdInfer    & 1.54          & 2.16          & 2.46          & 6.71          & 5.48          & 12.23          & 2.81          & 3.75          & 5.24          & 11.23          \\
CFR-1   & \textbf{1.44} & \textbf{2.04} & \textbf{2.35} & \textbf{6.43} & 5.33          & 11.99          & \textbf{2.70} & \textbf{3.58} & \textbf{5.11} & \textbf{11.14} \\
CFR-4   & 1.50          & 2.18          & 2.38          & 6.46          & 5.31          & 11.94          & 2.72          & 3.59          & 5.16          & 11.16          \\
A-CFR-4 & 1.50          & 2.06          & 2.39          & \textbf{6.43} & \textbf{5.28} & \textbf{11.90} & 2.73          & 3.60          & 5.17          & \textbf{11.14} \\
\hline 
\end{tabular}
\caption{\label{tab:casc} Results of CFR Schemes. The 'StdInfer' denotes the standard inference \cite{sofiiukRevivingIterativeTraining2022}; CFR-$n$ is CFR with fixed step $n$; and A-CFR-$n$ is the adaptive CFR with maximum steps $n$.}
\end{table*}

\begin{table*}[!th]
\centering
\small
\begin{tabular}{l|cc|cc|cc|cc|cc}
\hline
&
\multicolumn{2}{c}{GrabCut} & 
\multicolumn{2}{c}{PascalVOC} &
\multicolumn{2}{c}{SBD} &
\multicolumn{2}{c}{Berkeley} &
\multicolumn{2}{c}{DAVIS} \\
\hline
\multicolumn{1}{r|}{NoC@} & 90 & 95 & 90 & 95 & 90 & 95 & 90 & 95 & 90 & 95 \\
Model \\
\hline
\hline
!DIOS (FCN)                       & 6.04          & -             & 6.88          & -             & -             & -              & 8.65          & -             & -             & -             \\
!FCA-Net-SIS (Res2Net)            & 2.08          & -             & 2.69          & -             & -             & -              & 3.92          & -             & 7.57          & -             \\
\hline\hline                                                                                                                                                                                           
\#LD (VGG-19)                     & 4.79          & -             & -             & -             & 10.78         & -              & -             & -             & 9.57          & -             \\
\#BRS (DenseNet)                  & 3.60          & -             & -             & -             & 9.78          & -              & 5.08          & -             & 8.24          & -             \\
\#f-BRS-B (ResNet101)             & 2.60          & 4.82          & -             & -             & 7.05          & 13.81          & 4.13          & 10.05         & 7.31          & 14.30         \\
\#RITM (HRNet18)                  & 2.04          & 3.66          & -             & -             & 5.42          & 11.65          & 3.23          & 8.38          & 6.70          & 13.88         \\
\#CDNet (ResNet34)                & 2.64          & -             & -             & -             & 7.87          & -              & 3.69          & -             & 6.66          & -             \\
\#UCP-Net (EffNet)                & 2.76          & -             & -             & -             & -             & -              & 2.70          & -             & -             & -             \\
\#PseudoClick (HRNet18)           & 2.04          & -             & 2.74          & -             & 5.40          & -              & 3.23          & -             & 6.57          & -             \\
\#FocalClick (HRNet18)            & 2.06          & -             & -             & -             & 6.52          & -              & 3.14          & -             & 6.48          & -             \\
\#FocalClick (SegF-B0S2)          & 1.90          & -             & -             & -             & 6.51          & -              & 3.14          & -             & 7.06          & -             \\
\#SimpleClick (ViT-H $^{\text{Baseline}}$) & 1.44          & 2.10          & 2.20          & 3.01          & 4.15          & 9.86           & 2.09          & 6.34          & 5.33          & 11.67         \\
\hline
\#SimpleClick CFR-1 (ViT-H)               & \textbf{1.32} & 1.78          & \textbf{2.16} & 2.92          & \textbf{4.08} & \textbf{9.80}  & 2.15          & 6.21          & 5.22          & 11.64         \\
\#Ours (ViT-H)                            & 1.46          & 1.66          & 2.20          & 3.00          & 4.45          & 10.52          & 1.77          & 4.73          & 4.86          & 9.06          \\
\#Ours CFR-1 (ViT-H)                      & 1.42          & \textbf{1.62} & 2.17          & \textbf{2.91} & 4.45          & 10.50          & \textbf{1.74 $^{\text{-}16.7\%}$} & \textbf{4.44 $^{\text{-}30.0\%}$} & \textbf{4.77 $^{\text{-}10.5\%}$} & \textbf{8.85 $^{\text{-}24.2\%}$} \\
\hline\hline
-SAM (ViT-H)                      &1.84	          &2.18	          &2.72	       &3.71	       &7.59	       &14.97              &2.06	       &4.64	       &5.21	       &9.87	       \\
\hline
\hline
*RITM (HRNet18)                   & 1.54          & 2.22          & -             & -             & 6.05          & 12.47          & 2.26          & 6.46          & 5.74          & 12.45         \\
*RITM (HRNet32)                   & 1.56          & 2.48          & -             & -             & 5.71          & 12.00          & 2.10          & 5.41          & 5.34          & 11.52         \\
*PseudoClick (HRNet32)            & 1.50          & -             & 2.25          & -             & 5.54          & -              & 2.08          & -             & 5.11          & -             \\
*FocalClick (SegF-B3S2)           & 1.52          & 1.84          & 2.89          & 3.80          & 5.63          & 11.58          & 1.93          & 4.55          & 4.96          & 10.71         \\
*SimpleClick (ViT-H $^{\text{Baseline}}$)  & 1.50          & \textbf{1.66} & 1.98          & 2.51          & 4.70          & 10.76          & 1.75          & 4.34          & 4.78          & 8.88          \\
\hline
*SimpleClick CFR-1 (ViT-H)                & 1.56          & 1.76          & \textbf{1.94} & 2.46          & \textbf{4.60} & \textbf{10.74} & 1.67          & 4.20          & 4.72          & 8.76          \\
*Ours (ViT-H)                             & \textbf{1.48} & \textbf{1.66} & 1.99          & 2.52          & 4.81          & 10.94          & 1.51          & \textbf{2.90} & 4.27          & 7.62          \\
*Ours CFR-1 (ViT-H)                       & 1.58          & 1.76          & \textbf{1.94} & \textbf{2.45} & 4.74          & 10.90          & \textbf{1.46 $^{\text{-}16.6\%}$} & \textbf{2.90 $^{\text{-}33.2\%}$} & \textbf{4.24 $^{\text{-}11.3\%}$} & \textbf{7.50 $^{\text{-}15.6\%}$} \\
\hline
\end{tabular}
\caption{\label{tab:comp} Comparison with state-of-the-art approaches. '!' indicates a model trained on the Pascal VOC \cite{everinghamPascalVisualObject2010} dataset. '\#' denotes a model trained on the SBD \cite{hariharanSemanticContoursInverse2011} dataset. '-' denotes a model trained on the SA-1B \cite{kirillovSegmentAnything2023} dataset and, '*' denotes a model trained on the C+L \cite{sofiiukRevivingIterativeTraining2022} dataset. '-' represents unavailable value. Percentages that appear as superscripts indicate a reduction of number of clicks from the corresponding baseline model.}
\end{table*}

\section{Experiments}

\subsection{Datasets and Experiment Setup}

\textbf{Datasets.} The following image sets are used to evaluate the performance of our approaches.

\begin{itemize}
    \item \textbf{GrabCut \cite{rotherGrabCutInteractiveForeground2004}:} This comprises 50 images, each containing one instance.
    \item \textbf{Berkeley \cite{martinDatabaseHumanSegmented2001}:} This set consists of 96 images with 100 instances.
    \item \textbf{DAVIS \cite{perazziBenchmarkDatasetEvaluation2016}:} This dataset has 345 images extracted  by \cite{jangInteractiveImageSegmentation2019} from 50 videos with high-quality segmentation masks.
    \item \textbf{Pascal VOC \cite{everinghamPascalVisualObject2010}:} Only the validation set is utilized, featuring 1449 images and 3427 instances.
    \item \textbf{SBD \cite{hariharanSemanticContoursInverse2011}:} The training set comprises of 8497 images with 20172 instances. The validation set holds 2857 images with 6671 instances.
    \item \textbf{COCO \cite{linMicrosoftCocoCommon2014}+LVIS \cite{guptaLvisDatasetLarge2019} (C+L):} The COCO dataset involves 99k images with 1.2M instances in its training subset. The LVIS dataset includes 100k images with 1.2M total instances. A combined C+L \cite{sofiiukRevivingIterativeTraining2022} dataset was synthesized from COCO and LVIS encompassing 104k images and 1.6M instances, intended solely for training.
\end{itemize}

\textbf{Evaluation Metric}. We evaluate model performance using the Number of Clicks (NoC) metric. This standard measure quantifies the user inputs needed to achieve satisfactory segmentation results that surpass a pre-defined Intersection over Union (IoU) threshold. In this study, we report NoC@90 and Noc@95. Although earlier studies frequently reported NoC@85 and NoC@90 thresholds, the increasing demand for high-quality segmentation necessitates stricter criteria for model appraisal. We employ the method delineated in \cite{liInteractiveImageSegmentation2018} to generate clicks during evaluation. The maximum number of clicks is limited to 20.

\textbf{Backbone models}. In line with the methodology suggested by SimpleClick \cite{liuSimpleClickInteractiveImage2022}, we adopted the SimpleClick ViT-Base and ViT-Huge (ViT-B and ViT-H) models for our experiments. These models employ the Plain Vision Transformer \cite{dosovitskiyImageWorth16x162020a} as the backbone. Afterward, we fine-tuned these pre-trained models by leveraging our proposed methods. To further explore the adaptability of our methodological approaches, we also expanded our experiment to incorporate the HRNet \cite{sunDeepHighResolutionRepresentation2019} backbone.

\textbf{Implementation details.} Our Iterative Click Learning (ICL) utilizes three iteratively generated clicks for model training, setting $\beta_i \in [1, 2, 3]$. We employ the Adam optimizer, with parameters $\beta_1=0.9, \beta_2=0.999$, coupled with a learning rate of $5 \times 10^{-6}$. All models, fine-tuned over one epoch, are initially derived from the trained SimpleClick models. The batch sizes are set at 140 for Vit-B and 32 for Vit-H. We employ the Normalized Focal Loss (NFL) \cite{sofiiukAdaptisAdaptiveInstance2019} for training, with the parameters set at $\alpha=0.5$, $\gamma=2$. Beyond the proposed SUEM C\&P augmentation, we deploy image augmentations including resizing, flipping, rotation, brightness contrast adjustment, and cropping. The input images are unified to $448 \times 448$. The clicks are encoded into disk maps with a radius of 5. All models are trained on the NVIDIA Quadro RTX 8000 GPU.

\subsection{The Effectiveness of ICL and SUEM C\&P}

We fine-tuned the SimpleClick Vit-Base model, initially trained on the SBD dataset, using the proposed ICL and SUEM C\&P methods on the SBD dataset. Table \ref{tab:conv} compares the performance of the original SimpleClick model with the models fine-tuned via ICL and SUEM C\&P. As Table \ref{tab:conv} reveals, ICL enhances the SimpleClick model across all five datasets. In contrast, SUEM C\&P significantly advances the model's efficacy on the GrabCut, Berkeley, and DAVIS datasets, thus highlighting the benefits of ICL and SUEM C\&P.

To probe further into the applicability of the proposed training methods, we also applied ICL+SUEM C\&P on HRNet32 and compared it against the RITM and EMC-Click. Table \ref{tab:ritm} illustrates the results generated using ICL+SUEM C\&P with the HRNet32 backbone. These outcomes highlight a significant reduction of NoC@95 compared with the RITM model in the GrabCut, Berkeley, and DAVIS datasets by -20.97\%, -6.65\%, and -5.82\% respectively, when deploying the proposed method. The proposed method achieves competitive results on GrabCut, Pascal VOC, and SBD, and significantly outperforms EMC-Click on Berkeley and DAVIS datasets. This suggests the robust compatibility of ICL and SUEM C\&P with varying backbones.

\subsection{Inference Using the CFR Scheme}

Table \ref{tab:casc} validates the efficacy of the CFR inference process. We used both the fixed-step CFR and the adaptive CFR (A-CFR) for the inference phase of the SBD-trained SimpleClick ViT-B model, testing on the SBD test set. As shown in Table \ref{tab:casc}, all three CFR inference procedures (CFR-1, CFR-4, and A-CFR-4) outperform the standard inference method ('StdInfer') \cite{sofiiukRevivingIterativeTraining2022}, thus demonstrating CFR's capability to improve interactive segmentation performance.

Even though CFR-4 consumes more computational resource, its advancement over CFR-1 is not marginal, suggesting that increasing steps do not guarantee better outcomes. A-CFR-4 surpasses CFR-4 by adaptively halting refinement when changes fall below a set threshold, 20 pixels in this case. CFR-1 and A-CFR-4 show comparable performances but differ in aspects of efficiency and adaptability. For instance, CFR-1 performs only a single step, whereas A-CFR-4 can be fine-tuned with optimal thresholds to cater to specific situations. Following the assessment of computational efficiency, we selected CFR-1 for following experiments.

\subsection{Comparison With State-of-the-art}

Table \ref{tab:comp} details our proposed models' performances and twelve advanced deep learning-based interactive segmentation methods. Our approach ('Ours') incorporates ICL and SUEM C\&P to fine-tune the ViT-H model. In Table \ref{tab:comp}, separated by double solid lines and identified by symbols '!', '\#', '-', and '*', represent models trained on Pascal VOC, SBD, SA-1B \cite{kirillovSegmentAnything2023}, and C+L datasets, respectively. As Pascal VOC is small, its use in training is seldom reported.

Table \ref{tab:comp} demonstrates that SimpleClick ViT-H models trained on SBD and C+L exceed prior methods across five test datasets in NoC@90 and NoC@95. Consequently, we select these models as the baselines, denoted 'Baseline'.

For SBD trained models, applying our CFR-1 inference to the SimpleClick baseline model significantly decreases all metrics except NoC@90 on Berkeley. Specifically, CFR-1 decreases NoC@90 and NoC@95 on GrabCut by 8.3\%, and 15.2\% respectively. The proposed model, 'Ours CFR-1', built with ICL and SUEM C\&P, decreases clicks required for reaching an IoU of 0.9 (NoC@90) by 16.7\%. Importantly, it decreases NoC@95 by 30.0\% compared to the baseline, indicating effectiveness in high-precision image segmentation. Nevertheless, there is no remarkable improvement on the GrabCut, PascalVOC, and SBD testing sets. This can be attributed to the less complicated visual content in Grabcut and PascalVOC, and inaccurately annotated ground truths in SBD, rendering significant user click reduction with high IoU thresholds difficult.

Similar outcomes are observed for C+L dataset-trained models. The SimpleClick CFR-1 model improves NoC@90 and NoC@95 on all excluding GrabCut. Noteworthy improvements of the 'Ours CFR-1' model are visible on the Berkeley and DAVIS sets.

\section{Conclusion}
In this study, we present an interactive image segmentation framework based on deep neural networks (DNN), comprising three innovative components. These include: 1) iterative click loss (ICL)-based training, 2) cascade-forward refinement (CFR)-based inference, and 3) SUEM Copy-Paste image augmentation. The proposed framework is applicable to other iterative mask-guided interactive segmentation approaches. The proposed ICL introduces an innovative approach for minimizing the number of clicks. The CFR inference employs a unified framework that iteratively refines segmentation results using two loops. The proposed SUEM C\&P is a comprehensive image augmentation approach that generates more diverse training sets. Comprehensive experiments have been conducted to validate the effectiveness of all components, demonstrating that the proposed approach achieves state-of-the-art performance.

\bigskip
\printbibliography
\end{document}